\documentclass[jair-arxiv,twoside,11pt,theapa]{article}
\usepackage{jair-arxiv, theapa, rawfonts}

\jairheading{1}{1993}{1-1}{6/91}{9/91}
\ShortHeadings{Using Small MUSes to Explain How to Solve Pen and Paper Puzzles}
{Espasa, Gent, Hoffmann, Jefferson, Lynch, Salamon, McIlree}
\firstpageno{1}

\usepackage{soul}
\usepackage{url}
\usepackage[implicit=false]{hyperref}
\usepackage[utf8]{inputenc}
\usepackage{graphicx}
\usepackage{amsmath}
\usepackage{amssymb}
\usepackage{amsthm}
\usepackage{booktabs}
\usepackage{algorithm}
\usepackage{xspace}
\usepackage{siunitx}

\usepackage[noend]{algpseudocode}

\usepackage{listings}

\usepackage{tikz}
\usepackage[capitalise]{cleveref}
\usepackage{multirow}

\usepackage[a4paper,margin=3cm]{geometry}

\newcommand\removefornow[1]{}

\newcommand\demystify{\textsc{Demystify}\xspace}

\newcommand{\essence}[0]{\textsc{Essence}\xspace}

\newcommand{\savilerow}{\textsc{Savile Row}\xspace}

\definecolor{mygreen}{rgb}{0,0.6,0}
\lstdefinelanguage{essence}{
 frame = single,
 breaklines=true,
 keywords = { language, Essence, given, letting, find, such, that, domain, function, total, surjective, be , forAll, exists, injective, in, preImage, range ,  mset, set, partition, new, type, intersect, from , minimising, maximising, indexed, by, defined, maxSize, maxNumParts , subset , size, toInt, sum , sequence },
  keywordstyle=\color{blue}\bfseries,
  ndkeywords={atleast, atmost, gcc, int, matrix, bool},
  ndkeywordstyle=\color{mygreen}\bfseries,
  identifierstyle=\color{black},
  sensitive=false,
  comment=[l]{\$},
  commentstyle=\color{purple}\ttfamily,
  stringstyle=\color{red}\ttfamily,
  morestring=[b]',
  morestring=[b]",
}

\newcounter{row}
\newcounter{col}

\newcommand\setrow[9]{
	\setcounter{col}{1}
	\foreach \n in {#1, #2, #3, #4, #5, #6, #7, #8, #9} {
		\edef\x{\value{col} - 0.5}
		\edef\y{9.5 - \value{row}}
		\node[anchor=center] at (\x, \y) {\n};
		\stepcounter{col}
	}
	\stepcounter{row}
}

\newtheorem{definition}{Definition}

\title{Using Small MUSes to Explain How to Solve Pen and Paper Puzzles}

\author{\name Joan Espasa \email jea20@st-andrews.ac.uk\\
\name Ian P. Gent \email ian.gent@st-andrews.ac.uk\\
\name Ruth Hoffmann \email rh347@st-andrews.ac.uk\\
\name Christopher Jefferson \email caj21@st-andrews.ac.uk\\
\name Alice M. Lynch \email al254@st-andrews.ac.uk\\
\name Andr\'as Salamon \email andras.salamon@st-andrews.ac.uk\\
\addr University of St Andrews \\
\name Matthew J. McIlree \email m.mcilree.1@research.gla.ac.uk \\
\addr University of Glasgow
}

\begin{document}

\maketitle

\begin{abstract}

In this paper, we present \demystify, a general tool for creating human-interpretable step-by-step explanations of how to solve a wide range of pen and paper puzzles from a high-level logical description.  \demystify is based on Minimal Unsatisfiable Subsets (MUSes), which allow \demystify to solve puzzles as a series of logical deductions by identifying which parts of the puzzle are required to progress. 
This paper makes three contributions over previous work.  First, we provide a generic input language, based on the \essence constraint language, which allows us to easily use MUSes to solve a much wider range of pen and paper puzzles.
Second, we demonstrate that the explanations that \demystify produces match those provided by humans by comparing our results with those provided independently by puzzle experts on a range of puzzles.
We compare \demystify to published guides for solving a range of different pen and paper puzzles and show that by using MUSes, \demystify produces solving strategies which closely match human-produced guides to solving those same puzzles (on average 89\% of the time). 
Finally, we introduce a new randomised algorithm to find MUSes for more difficult puzzles. This algorithm is focused on optimised search for individual small MUSes.

\end{abstract}

\section{Introduction}

``Pen and paper" logic puzzles are puzzles designed to be solved on paper -- they appear in newspaper pages, magazines and specialist books. Popular pen and paper logic puzzles include Sudoku, Futoshiki and Skyscrapers\footnote{In this paper we will not consider puzzles based around language, such as Crosswords.}, but new puzzles and variants are created almost weekly, and there are many websites and books dedicated to showing off new problems. 
Pen and paper puzzles are also popular on the internet, for example the YouTube channel `Cracking the Cryptic' explain how to solve a range of unusual and difficult pen and paper puzzles~\cite{miracle}.

While many pen and paper puzzles are NP-complete \cite{Kendall2008}, instances intended to be solved by humans can usually be solved quickly (for example, Sudoku with a constraint solver \cite{simonis2005sudoku}).
Most Artificial Intelligence (AI) systems use a branching and backtracking search, while human players try to solve puzzles using only deductions -- many players believe they should never guess then backtrack \shortcite{DBLP:conf/cp/HoffmannZAN22}.
Constraint solvers use reasoning algorithms known as propagators along with backtrack search. There are two issues with using constraint solvers to try to guide human players.
If we use weak propagators, we produce search trees with hundreds or thousands of nodes, which are too large to be understood by human players.
If we use strong propagators or algorithms like Singleton Arc Consistency~\cite{DebruyneB97} we can make deductions beyond the abilities of most human players while still usually producing search trees. In contrast, human players aim to solve problems with no backtrack.

There are two main reasons to look at how humans solve puzzles -- to advise players on how to progress and to produce more accurate difficulty measures of puzzles. 
For example,~\citeA{Pelanek2011DifficultyModel} measures the difficulty of Sudoku by looking at the number and difficulty of deductions which can be applied at each point in solving. \citeauthor{Pelanek2011DifficultyModel} uses a hard-wired model of possible deductions for Sudoku created from the existing Sudoku literature. %
A common approach to creating solvers which can explain how a puzzle is solved is to create custom solvers which use the same techniques as human players. Building such a solver requires a puzzle-specific list of techniques which human players use.
For popular puzzles the techniques which human players use are well documented, often in the form of guides which teach other players~\cite{conceptis,tectonic,Stuart2008SudokuWiki,Wilson2006HowGuide}. These guides often also cover popular variants of the most famous puzzles. 
SudokuWiki~\cite{Stuart2008SudokuWiki} is an example of a website with solvers for several popular Sudoku variants, showing which techniques can be applied at each stage of solving. The major limitation of these systems is the requirement for a hard-wired and ordered list of techniques -- even slightly different puzzles usually require a different variant solver.

We build on the work of~\citeA{bogaerts2020step} in using MUSes (Minimal Unsatisfiable Subsets) as a basis for providing interpretable explanations of puzzles from a SAT model.  While~\citeauthor{bogaerts2020step} consider using MUSes as a technique for solving puzzles in a human-like way, they considered Logic Grid Puzzles (which only require a Boolean domain), and the results were not compared to human players.
When considering explanations, interpretability is defined as ``descriptions that are simple enough for a person to understand, using vocabulary meaningful to the user''~\shortcite{GilpinBYBSK18}. 

This paper makes three contributions over previous work.  First, we provide a generic input language, based on the \essence constraint language, which allows us to easily use MUSes to solve a much wider range of pen and paper puzzles.

Second, we demonstrate that the explanations that \demystify produces match those provided by humans by comparing our results with those provided independently by puzzle experts on a range of puzzles.

Finally, we introduce a new randomised algorithm to find MUSes for more difficult puzzles. This algorithm is focused on optimised search for individual small MUSes.

\section{Background}

Most puzzles we discuss in this paper require filling in a set of boxes (usually called \emph{cells}) with a number or symbol in each cell (called \emph{values}), such that a set of constraints are all satisfied. We keep a list of the values being considered for each cell, called the \emph{candidates}. Solving consists of a series of logical deductions, each of which remove one or more candidates from cells which we can prove are not part of the solution to the puzzle. Once every cell has only one candidate remaining, and the rules/constraints of the puzzle are not violated, the puzzle is solved. 

We consider puzzles which have only a single solution, can be checked for correctness by players using only the provided rules, and are intended to be solved by humans without the need to ever guess. We call these \emph{pen and paper puzzles}. 

The fact that puzzles must have a single solution is a natural consequence of puzzles which can be solved without guessing -- in a puzzle with more than one solution, at least one cell must be able to take two different values in different solutions. This means it is impossible to deduce a single value for every cell purely using logic, the player must at some point make a choice on what value to assign to a cell.

\subsection{Automated versus Human Solving of Puzzles}

Automated solving of puzzle games by AI, when focusing on the speed or efficiency rather than mimicking people, uses propagators and search trees~\cite{simonis2005sudoku}. 
Some puzzle-solving guides aimed at people include small search trees, but these are considered a last resort option~\cite{Stuart2008SudokuWiki}. Guides aimed at people instead encourage tracking possible values, logical deduction, and pattern recognition~\cite{Stuart2008SudokuWiki}.

There have been various attempts to produce systems that mimic human solving approaches. These generally define a heuristic to find the simplest deduction to be made next in a given puzzle, and then applying that deduction (or a randomly selected deduction if several of equal simplicity are available). Previous attempts have been puzzle specific, such as a program that can solve a Sudoku, and the heuristics employed vary in sophistication. A common approach is to implement a set of strategies described by puzzle solving guides, and use the difficulty ordering specified by these guides as the measure of difficulty~\cite{Pelanek2014DifficultyEvaluation}. 

\citeauthor{bogaerts2020step} produced explanations of human-possible deductions for grid-based logic puzzles, using a constraint solver.
Their approach, like ours, involves identifying small explanations which are supposed to be similar to those provided by humans~\cite{bogaerts2020step}.
However, our approach is randomized rather than exhaustive and applicable to non-Boolean value domains.
Randomization allows us to deal with many puzzles that are too difficult for an approach relying on solving the underlying difficult optimisation problem.
Further, we provide evidence that our explanations correspond well to how humans approach solving puzzles, by showing how these explanations correspond to the approach taken in tutorials produced by expert human puzzlers.

 \subsection{SAT, Constraints, and Automated Modelling}
 
Boolean satisfiability (SAT) is a widely used formalism for expressing and solving combinatorial search problems, which searches for satisfying assignments to a Boolean formula expressed in conjunctive normal form~\cite{handbooksat}.
Constraint programming (CP) provides a more expressive language than that used in SAT, featuring arithmetic and high-level operations.
A CP problem consists of a list of \emph{variables}, each with a list of \emph{values} it can take, and a list of \emph{constraints} applied to subsets of the variables.
A \emph{solution} is an assignment of one value to each variable, such that each constraint is satisfied by the solution~\cite{handbookcp}.
SAT can be regarded as a special case of CP with all variables taking only one of two possible values, but in practice these methods use different techniques to obtain solutions, and a common technique for solving CP problems is to use a tool which converts them into a SAT problem.
CP allows constraints to succinctly express what might require many SAT clauses.

\subsection{Minimal Unsatisfiable Sets} \label{sec:backgroundmus}
\begin{definition}[Unsatisfiable Set/Core]
An \emph{unsatisfiable set} of an unsatisfiable constraint problem is any unsatisfiable subset of the set of constraints of the problem.
\end{definition}

Note that unsatisfiable sets can be only be found in problems which are already unsolvable. In puzzles we will use unsatisfiable sets to solve ``what if'' problems by forcing a cell to take a value which it does \textbf{not} take in the solution to the puzzle, then find unsatisfiable sets in this unsolvable problem.

Traditionally, unsatisfiable sets are defined on the clauses of a conjunctive normal formula. We extend this definition to general constraint problems. A given problem can have many unsatisfiable sets of different sizes.
The main hypothesis we test in this paper is that unsatisfiable sets closely align with how human players solve puzzles on a wide range of puzzles.
Unsatisfiable sets have many other uses, such as interactive applications~\cite{junker2001quickxplain}, repairing 
knowledge bases~\cite{muskb} or model checking~\cite{musmodelchecking}, and extensive surveys are available~\cite{mussurvey1,mussurvey2}.

Identifying \emph{minimum} unsatisfiable sets is a $\sum_{2}$-complete problem~\cite{muscomplexity}.
On the other hand, the decision problem of a 
\emph{minimal} (i.e.~irreducible) unsatisfiable set can be formulated as the difference between two NP problems,
which lies in the DP-complete~\cite{PapadimitriouW88} class.  
Different approaches have been studied for computing \emph{Minimal Unsatisfiable Sets} (\emph{MUS}) such as
insertion-based~\cite{NP88}, deletion-based~\cite{ChinneckD91}, dichotomic~\cite{HemeryLSB06} or by 
progression~\cite{Marques-SilvaJB13}.
As most techniques that require unsatisfiable sets do not strictly require them to be minimum, most specialised
tools such as Muser2~\cite{BelovM12}, HaifaMUC~\cite{NadelRS13}, MCS-MUS~\cite{BacchusK15}, 
TarmoMUS~\cite{WieringaH13} or some modern SAT solvers try to find minimal or even small enough unsatisfiable sets, striking a balance between size and practicality. We investigated these systems but found they could not produce good MUSes for the puzzles we considered, which is why we develop the randomised algorithm given in this paper.

\subsection{Explanations as Minimal Unsatisfiable Subsets}

This section discusses using MUSes as the basis for human-understandable explanations, with the aim to understand how they both match and how they differ. We argue that MUSes can reasonably be seen as explanations in our context and also discuss their limitations.

The work of~\citeA{bogaerts2020step} already uses MUSes as a basis for providing interpretable explanations, stating that the interpretability of an explanation is not only problem-specific but also subjective. Their proposed framework is general enough to allow a problem-specific cost function that quantifies the interpretability of a single explanation. This framework is then evaluated using a size-based metric on the extracted MUSes.

We consider each puzzle in Section~\ref{sec:puzzles} as a Constraint Satisfaction Problem (CSP), containing a set of variables and a set of constraints. The variables represent the cells which the player must find the values for, while the constraints model the puzzle rules. %

We assume that players will generally solve puzzles as a series of steps. At each step, they will deduce some new information, which can be represented by removing one or more possible values from the domain of one or more variables. 
At its most abstract, we will consider an \emph{explanation} a logical argument that proves that some set of variable-value pairs does not occur in a solution to the puzzle. Further, if we can deduce the assignment for a variable, that is equivalent to proving it is not able to take any other value in its domain.

As noted in~\citeA{bogaerts2020step}, for an explanation to be easily comprehended by a player, it should ideally be small, only involving a few constraints of the problem.
In most cases, when explaining why a variable in a puzzle cannot be assigned a value, only a small subset of the constraints will be needed. This requirement aligns well with the definition of a MUS, which minimises the number of constraints that makes a problem unsatisfiable. Therefore, we can get an explanation for why variable $x$ cannot be assigned value $v$ by stating $x = v$ and then asking the solver for a MUS.

\subsection{Pen and Paper Puzzles Considered In This Paper}\label{sec:puzzles}

There are a wide range of pen and paper puzzles. In the experiments in this paper we use the following puzzles. 
There is an extensive selection of available pen and paper puzzles, with many variations for each one of them. As it would be unfeasible to study each one of them, a carefully chosen set of puzzles follows, selected for their coverage of the different constraints found in most well-known puzzles.

For Sudoku, X-Sudoku and Jigsaw Sudoku we used~\citeS{Stuart2008SudokuWiki} SudokuWiki.
The other two major sources for various puzzles, are~\citeA{conceptis}
and~\citeA{tectonic}.

\emph{Binairo}, also known as Takuzu, is a partially completed grid of 0s and 1s where empty cells must also be $0$ or $1$ such that: each row and column has an equal number of $0$s and $1$s, all rows or columns are different and there cannot be three $0$s or three $1$s in a continuous sequence in any row or column.

\emph{Futoshiki} is a puzzle with an $n\times n$ grid, containing inequality constraints between some pairs of adjacent cells. 
The aim is to fill in the cells with numbers from $1$ to $n$ so that each row and column contains each possible value exactly once and all inequalities are satisfied.

\emph{Kakuro} is a crossword-style puzzle with numbers instead of letters.
Each cell contains a digit from $1$ to $9$, such that the cells add up to the hint given for that sub-row (or sub-column). Also, each sum consists of unique digits.

\emph{Skyscrapers} is an $n\times n$ grid based puzzle where each cell contains a number from $1$ to $n$. These numbers are interpreted as the heights of skyscrapers, with $1$ being the lowest and $n$ being the highest. The grid represents a top-down view of the skyscrapers.
Each number must occur exactly once in each row or each column. Around the outside of the grid there is a number which represents how many skyscrapers could be ``seen'' from this location. 

\emph{Sudoku} is a logic puzzle with the goal of filling a (usually) \(9 \times 9\) grid of cells, where each cell contains a number between \(1\) and \(9\), so that each column, row, and defined \(3\times 3\) box (usually marked with a bold outline) contain each of the numbers \(1\) to \(9\) exactly once. 
The grid contains some initial values and has a single solution.
The popularity of Sudoku has led to many variants (such as the Jigsaw Sudoku, X-Sudoku and the Miracle Sudoku). 
These variants keep an \(n \times n\) grid where each row and column contains the numbers from \(1\) to \(n\), but may vary the shape of the boxes, or add extra constraints. 

\emph{Starbattle} is a puzzle on an $n \times n$ grid with a number of marked regions (each region consisting of a connected set of cells). The puzzle requires each row, column and region to contain the same number of stars. 
The number of stars can range from $1$ to $n$, each puzzle indicates 
how many stars are required. Stars can also not be placed in adjacent cells (orthogonally or diagonally).

\emph{Tents and Trees} has a grid which begins with some cells containing a tree, and for each row and column the number of tents which should appear in it.
The goal is to fill in the remaining cells with either a tent or grass, such that each tree is paired with an orthogonally adjacent tent, the number of tents reflect the expected number for each row and column, and no two tents are touching (orthogonally and diagonally).

\emph{Thermometer} is a grid filled with thermometers of different lengths, which are each place in a sequence of either vertical or horizontal cells, such that every cell is part of some thermometer. The goal is to partially fill each thermometer with mercury from the ``bottom'' (a cell with a rounded end).  Thermometers do not have to be filled to the top.
The hints on the outside edges of the grid indicate how many mercury filled cells there are in the corresponding column or row.

\section{\demystify}
The \demystify \footnote{https://github.com/stacs-cp/demystify} system can be broken into three parts, which we will discuss in this section.

\begin{enumerate}
    \item A high-level declarative input language, that is an extension to the \essence constraint language. This language allows us to easily express a wide range of puzzles, and provide human-readable descriptions of each constraint of the puzzle.
    \item A Python library which uses a randomised MUS finding algorithm to probuce a step by step guide to solving a given puzzle. 
    \item An extensible web-based frontend which displays a wide range of puzzles. This interface shows explanations, highlighting the location of filled in values and the exact location of the constraints used in the explanation. By default the frontend solves a complete puzzle in the most efficient way and shows the solution step by step, users can also choose the order in which the puzzle is solved, using an explanation for each step in the puzzle solving process.
\end{enumerate}

\subsection{Finding Step by Step Explanations}\label{sec:stepbystep}

\demystify generates explanations in a similar style to~\citeA{bogaerts2020step}.
Below is a schematic description showing our general procedure of explaining decisions when solving a puzzle. We start with a description of the puzzle rules.

\begin{enumerate}
    \item Translate the puzzle constraints to a CNF (Conjunctive Normal Form) formula $P$ using SavileRow~\cite{SR}. This translation produces, amongst other things, a set of Boolean variables $L$, one for each value which can be assigned to each \textit{problem} variable. There is also the set $X$ of Booleans, representing the \textit{constraint} variables of the puzzle. Each $x \in X$ is associated with a constraint of the puzzle $c$, where $x \rightarrow c$. We will go into more detail in Section~\ref{modelling}.
    \item For each $l \in L$ take its value $a$ in the unique solution and find MUSes for the problem $P \wedge (l \neq a)$.
    \item Pick a variable \(l \in L\) with the best MUS $M$ (under some metric) and display this to the user.
    \item Assign any literals for which $M$ is a MUS (which will include $l$, but may include others) and iterate from step 2 until all variables are assigned.\footnote{A `literal' is either a negated or non-negated propositional variable in the CNF formula.}
\end{enumerate}

\subsection{The \demystify Input Language}\label{modelling}

The input language of \demystify is an extension of the \essence modelling language \cite{conjurejournal}. \essence is a general-purpose high-level modelling language used to express and solve a large range of industrial and academic problems.

For the purposes of this paper, an \essence specification of a problem can be divided into three sections:

\begin{enumerate}
    \item The \texttt{given} statements specify any inputs of the problem. These allow a single \essence specification to represent a class of problems (such as all Sudoku problems or all Tents and Trees problems), with the \texttt{given} statements representing a specific instance of the problem, such as the partially filled in starting grid of a puzzle.
    \item The \texttt{find} statements represent the finite domain variables whose values are to be found. Within the problems we consider here, these will always be Booleans, integers of a finite domain, and matrices of Booleans or integers.
    \item The \texttt{such that} statements represent the constraints of the problem, which place conditions on values taken by the variables defined earlier -- any solution must satisfy every constraint.
\end{enumerate}

The \essence language can already express all of the puzzles we are interested in, as well as having other features not discussed or required in this paper. A full description of \essence can be found elsewhere \cite{frisch2008essence}.
The extensions listed below allow us to annotate the variables (as given in the \texttt{find} statements) with their purpose in the puzzle and provide human-readable descriptions of their purpose (both the cells, and the constraints). There are three types of annotations and every variable must have exactly one annotation.

\begin{itemize}
    \item[\texttt{VAR}]: variables (or matrices of variables) for which the puzzle player must find values.
    \item[\texttt{CON}]: variables (or matrices of variables) representing a constraint of the problem. These are described in more detail below.
    \item[\texttt{AUX}]: variables (or matrices of variables) that aid in expressing the puzzle in \essence, and which should be invisible to the player. This last category could be deduced from the lack of a \texttt{VAR} or \texttt{CON} annotation, but we require it so modellers consider the annotation of all variables.
\end{itemize}

Part of an example model is given in Figure \ref{fig:partsudoku}, it is a model of the Sudoku rules. Lines beginning \texttt{\$\#} denote \demystify annotations. As \texttt{\$} starts a comment in \essence, all \demystify inputs are also valid \essence specifications. \texttt{fixed} specifies the initially known values of the Sudoku (where 0 represents a value to be filled in by the user). \texttt{grid} is the variables which the user will complete.
    
The matrix of variables \texttt{con\_ad} represents the constraint that values in a row must be all different -- the variable \texttt{con\_ad[i,j1,j2,d]} represents that \texttt{grid[i,j1]} and \texttt{grid[i,j2]} cannot both be \texttt{d}. The string at the end of the \texttt{\#\$CON} gives an English description of the constraint. This notation allows arbitrary Python statements inside \(\{\}\) -- the array \texttt{a} is the \textit{assignment} of the indices of the variable. Any \texttt{given} variables can also be referred to by name.

The variables in a \texttt{CON} should always be used in the form \(v \implies con\). These implications allow our MUS finding algorithms to force \(con\) to be true by setting \(v\) to true. Note that we do not use \(v \iff con\) because when a MUS does not contain \(con\), this does not mean \(con\) has to be false. We call these \texttt{CON} variables \emph{activation variables}.

Some existing MUS finding systems to not require explicit annotations \cite{10.1007/978-3-319-59776-8_7}. However, we require constraint annotations for two reasons.  Firstly it gives us a way to attach a human-understandable explanation to each constraint, and secondly it lets us pick exactly which expressions are considered when building the MUS -- automatic systems can consider only part of a constraint, which may not map to a sensible human-interpretable expression.

\begin{figure}
    \centering
    \begin{lstlisting}[escapechar=|, language=essence]
letting D be domain int(1..9)
given fixed : matrix indexed by [D,D] of int(0..9)

$#VAR grid
find grid : matrix indexed by [D,D] of D

such that
forAll i,j: D.
    fixed[i,j] != 0 -> grid[i,j]=fixed[i,j],

$#CON con_ad "cells ({a[0]},{a[1]}) and ({a[0]},{a[2]}) cannot both be {a[3]} as they are in the same column"
find con_ad: matrix indexed by [D,D,D,D] of bool

such that
forAll i:D.
    forAll j1,j2:D. j1 < j2 ->
        forAll d:D. con_ad[i,j1,j2,d] -> !(grid[i,j1] = d /\ grid[i,j2]=d),

    \end{lstlisting}
    \caption{Part of a Sudoku model in \demystify}
    \label{fig:partsudoku}
\end{figure}

\subsection{Modelling Puzzles in \demystify}

\demystify will accept any \essence model, assuming it has been labelled with at least \texttt{VAR} and \texttt{CON} annotations and there is only one solution to the \texttt{VAR} variables when all \texttt{CON} variables are true.
When modelling puzzles, we first produced a standard \essence model of the puzzle and then attached a \texttt{CON} variable (with natural language description) to each constraint. Where constraints are quantified with \texttt{forAll}, we label each iteration of the \texttt{forAll} separately. In some cases, discussed below, we split a single constraint into multiple constraints to create finer-grained explanations. The \essence models of all puzzles we consider are contained in the \demystify repository.
We describe the most significant representation choices in the rest of this section, most notably the \textit{AllDifferent} constraint.

\subsubsection{The AllDifferent Constraint}

The \textit{AllDifferent} constraint constrains a list of variables \(X\) to all take different values. The \textit{AllDifferent} constraint occurs naturally in many problems and has been extensively studied \cite{PeteAllDiff}. The constraint occurs naturally in many pen and paper puzzles, most notably Sudoku, which consists of only \textit{AllDifferent} constraints.

In most puzzles, \textit{AllDifferent} constraints are imposed on \(n\) variables, each of which can take the same \(n\) values. In this situation, not only do the variables all have to take different values, but also every value must be assigned to some variable. We only consider these types of \textit{AllDifferent} constraints in this section. In particular, we will discuss how this looks in the context of Sudoku puzzles, but similar reasoning can be applied to for example Kakuro puzzles.

We will begin by considering all possible reasoning which can be performed given 
a single \textit{AllDifferent} constraint. This reasoning has been extensively studied by the
Sudoku community. One family of Sudoku techniques, known as the \textit{naked} reasoning rules, consist of the \textit{naked single, naked pair, naked triple} and \textit{naked quad} techniques.
This family of techniques reasons that if there are \(n\) cells which together only have 
\(n\) values they can take, then those \(n\) values must be taken by these \(n\) cells. This reasoning implies
those values can be eliminated from every other cell in the \textit{AllDifferent}. Similarly,
there are also the \textit{hidden single, hidden pair, hidden triple} and \textit{hidden quad} techniques, which
say that if there are \(n\) values only allowed by \(n\) cells, then those \(n\) cells must
take those values, so every other value can be removed from those cells. Note that only the
\textit{hidden} techniques require that our \textit{AllDifferent} constraint has the same
number of cells as possible values. For example if we required that 3 variables with domain \(\{1,\ldots,9\}\)
were \textit{AllDifferent}, we could not consider \textit{hidden singles}, as just because only one of those three variables could take the value \(7\) would not mean it had to be \(7\).

The problem of what reasoning can be performed with a single \textit{AllDifferent} constraint is well-studied, and involves a concept known as Hall Sets.
A Hall Set~\cite{viennot1974algebres} is a set of variables \(S\)
where the number of values which can be taken by variables in \(S\) is the same size as \(S\) -- the same as the \textit{naked} techniques described above.
The \textit{hidden} techniques are also Hall Sets, just of larger sizes.
For example, a \textit{hidden single} requires there is a value which occurs in only one cell, which
is the same as saying the other 8 cells together only take 8 values, so we could call this a
\textit{``naked eight''}. 
The named \textit{hidden} and \textit{naked} techniques describe every possible size of Hall Set, except the trivial case of a single variable having a single allowed value, so capture all possible reasoning.

We could represent \textit{AllDifferent} constraints as a single constraint when finding
MUSes, but this has a significant weakness. 
It would make all \textit{naked} and \textit{hidden}
techniques have the same complexity (a MUS of size 1) while larger
\textit{naked} and \textit{hidden} techniques are considered more complicated \cite{stuart2007sudoku}
. %

We, therefore, break each \textit{AllDifferent} constraint into smaller constraints.
As a first step, we break it into constraints of the form \(x_i \neq k \vee x_j \neq k\) 
for \(i \neq j\). We describe this as ``$x_i$ and $x_j$ cannot both be $k$ as they must be different''. This produces MUSes of reasonable size for many Sudoku techniques.

However, one significant limitation of this formulation is that the \textit{hidden single},
generally considered an easy technique, requires an extremely large MUS. Therefore we add
another constraint, \((\sum_i x_i = k) \geq 1\), described as ``at least one cell in the \textit{AllDifferent} must take the value $k$''. While this could be tightened to exactly 1, we tried to keep each constraint as weak as possible, as we found this made the resulting MUSes easier to interpret.

Similarly to how there is an extensive literature in CP and SAT on how to best represent
global constraints such as \textit{AllDifferent} most effectively, we expect there will
be similar research into how to best represent constraints for explanations and this
will be an important part of future work in this area.

\subsubsection{Counting Constraints}

A number of problems involve counting the number of objects which satisfy some condition, for example Tents and Trees and Thermometers can both give a ``sum" for each row and column, which gives the number of objects that appear.

Modelling each sum as a single constraint works correctly, but we noticed in descriptions of how such puzzles are solved it is common to need only one of ``the row contains at least \(X\) objects'' or ``the row contains at most \(X\) objects'', but rarely the exact sum. We therefore break such sums into a \(\leq\) and a \(\geq\). Of course a MUS can still contain both parts of the sum constraint, which together imply the sum of the row/column is \(X\).

\subsubsection{Pairing Constraint}

In the ``Tents and Trees'' puzzle, each tree must be paired with a tent. Ensuring there there is a one-to-one mapping between trees and tents is difficult to write in a way where it produces good explanations. We then label each tree with a distinct negative integers, and tents with distinct positive integers, requiring that tree $-x$ is paired with tent $x$.

\subsection{Refining \essence to SAT}
As previously discussed, given an augmented \essence model, Conjure and SavileRow are used to refine the model into SAT. For a full discussion of how Conjure and SavileRow operate, see \cite{conjurejournal}, this paper only discusses \demystify{}-specific extensions. Sudoku (see \Cref{fig:partsudoku}) is used as an example throughout this subsection.

\subsubsection{Identification of variables}

When refining \essence to SAT, many SAT variables are created. For an integer variable \(x\) with domain \(D\), SAT variables which represent \(x=d\) and \(x \leq d\) for each \(d \in D\) may be created. These SAT variables are usually only created when they are required to produce a correct and efficient model. We extend \savilerow to always create the variables which represent \(x=d\) and output a mapping which describes how the domain of each variable is represented in SAT.

\subsubsection{Disabling Optimisations}

Traditionally the SAT instances outputted by \savilerow are only intended to find one, or all, solutions to a problem. \demystify takes the generated SAT models and uses them to solve many subproblems, by assigning subsets of the variales, and in some cases this is not compatable with the optimisations \savilerow can use. As an example, \savilerow adds constraints to break symmetry of problems. While this allows problems to be solved faster, symmetry breaking constraints can lead to invalid MUSes. We checked the optimisations performed by \savilerow and disabled any which could affect the MUSes of the generated SAT instances -- these all involved symmetry breaking.

\subsubsection{Removal of trivial constraints}

Because matrices of variables in \essence are square, it is common for some \texttt{CON} variables to not be used in any constraint. For example, in Sudoku we use \texttt{con\_ad[a,b,c,d]} to represent that locations \texttt{(a,b)} and \texttt{(a,c)} cannot both take value \texttt{d}. This constraint is not generated when \texttt{b=c}, as in that case it would not make sense. This means that \texttt{con\_ad[a,b,c,d]} is not used in any constraint when \texttt{b=c}.

Rather than require users to explicitly list which of the \texttt{con\_ad} variables are used for constraints, we instead check if any \texttt{CON} variable is not contained in any SAT clause. Such variables can never appear in a MUS, as changing their value cannot affect if the problem is solvable, so we remove these constraints.

In some cases it is immediately obvious a \texttt{CON} variable does not take part in a constraint, for example \texttt{con\_ad[1,1,1,1]}. In other cases this may be less obvious -- for example if a Sudoku initially has \(5\) filled in the top right box, then \texttt{con\_ad[1,1,2,8]} (cells \((1,1)\) and \((1,2)\) cannot both be \(8\)) will be trivially satisfied, so \savilerow will not output any constraints on this variable.

\subsubsection{Removal of identical constraints}

In some puzzles, the natural method of expressing the constraints can generate identical constraints. For example, in Sudoku there are two reasons that cells \((1,1)\) and \((1,2)\) cannot both be \(5\) -- because the cells are in the same row and also because they are in the top-left 3x3 box. Having identical constraints like these is not incorrect but creates unnecessary extra work. No MUS would contain both of these constraints (as then it would not be minimal), and any MUS with one constraint could contain the other equally well.

In general, it is NP-hard to prove that two constraints are identical, but in our experiments \demystify was able to find all identical constraints that we noticed in the puzzles we considered. \demystify finds identical constraints by looking at the SAT instance.  For each \texttt{CON} activation variable \(b\), it gathers the SAT clauses which involve each constraint activation variable \(b\). If we find the set of clauses two activation variables are identical (other than the activation variable itself), then we know these variables represent identical constraints and one can be eliminated. One reason this method is successful in practice is that \savilerow will merge identical constraints, as it features common sub-expression elimination.

\subsubsection{Detection of constraint scope}

When visualising constraints, the \demystify visualiser highlights the literals in each constraint's scope. 
The scope of a constraint is traditionally the variables that it refers to. We use a non-standard definition as we want to highlight the smallest possible set of literals. There are two reasons in practice we want to use a non-standard definition:

\begin{enumerate}
    \item We want (as far as possible) a \emph{minimal} scope, for example the scope of the constraint \(2*a+0*b+1*c \geq 0\) is just the variables \(a\) and \(c\). This type of expression, where the values of some variables do not have an effect on if the constraint is satisfied, appear in several puzzles we consider.
    \item Given a constraint such as \((x\neq 2) \vee (y \neq 2)\), we wish to only highlight the literals \((x,2)\) and \((y,2)\), as checking if this constraint is true requires only knowing if these two literals are in the solution.
\end{enumerate}

We define a (not necessarily minimal) \emph{scope} of a constraint \(c\) to be a set of literals \(L\) such that knowing which literals in \(L\) are part of the assignment is sufficient to know if the constraint is true or false.

Finding a minimal scope of a constraint is non-trivial. We use the SAT output of SavileRow to calculate the scope. Starting at the activation variable for a constraint, we perform a search over the SAT instance, looking for all literals involved in the constraint. This algorithm is performed recursively:

\begin{itemize}
    \item If a SAT literal \(x\) is in the scope, and SavileRow has denoted that \(x\) represents a problem variable \(p\) being assigned a value \(c\), then \(p=c\) is in the scope.
    \item If a SAT literal \(x\) is in the scope, and SavileRow has denoted that \(x\) represents a problem variable \(p\) being assigned a value less than or equal \(c\), then all assignments to \(x\) less than or equal to \(c\) are in the scope.
    \item Otherwise, if a SAT literal \(x\) is in the scope, then all literals in all clauses containing $\neg x$ are in the scope.
\end{itemize}

\subsection{MUS selection}\label{sec:musselect}

There are a number of different ways of selecting which MUS is the ``best'' to show to users. In general, we have a small set of goals (which may conflict)

\begin{itemize}
    \item MUSes which involve the fewest constraints.
    \item MUSes which involve the fewest literals.
    \item MUSes which make the most progress towards solving the problem.
\end{itemize} 

The first of these goals is the simplest -- in general a MUS which involves fewer constraints will be easier to interpret. This does assume that all constraints are equally difficult to interpret, while it would be simple to extend our MUS finding algorithms to allow weighting constraints, we leave this for future work.

The second goal minimises the number of literals that have to be considered. Given the first goal, this will both consider smaller constraints, and consider constraints with overlapping scopes.

The final goal is implemented in an additional step which tries to use the same MUS to derive multiple literals. Whenever we find a MUS for a given literal, \demystify looks at all the literals involved in the MUS and checks if this is also a valid MUS for those literals. For example, consider that the single constraint \(A+B+C \leq 1\) with \(A,B \in \{0,1\}\) and \(C \in \{1\}\), was found as a MUS for \(A=1\). After checking each literal in the scope of the constraint, we find that this constraint is also a MUS for \(B=1\). Reusing MUSes to eliminate multiple literals both reduces the number of steps, and amount of repetition required to solve the puzzle.

Our criteria for picking the ``best'' MUS uses the following ordering, which aligns with our goals from above:

\begin{itemize}
    \item First choose MUSes with the fewest constraints
    \item Break ties by choosing the MUSes which involve the fewest literals in the union of the scopes of the constraints in the MUS
    \item Choose the MUSes which eliminate the most literals.
    \item Finally, order MUSes using the natural language description of the constraints in the MUS under lexicograpic ordering. This final step does not produce better MUSes, but ensures \demystify produces repeatable output.
\end{itemize}

The number of constraints and number of literals in a MUS can be seen as a proxy for the `difficulty' of the MUS. In future work we plan to build better models for the relative difficulty of MUSes.

\subsection{Visualisation}

\begin{figure}[ht]
    \begin{center}
        \includegraphics[width=\textwidth]{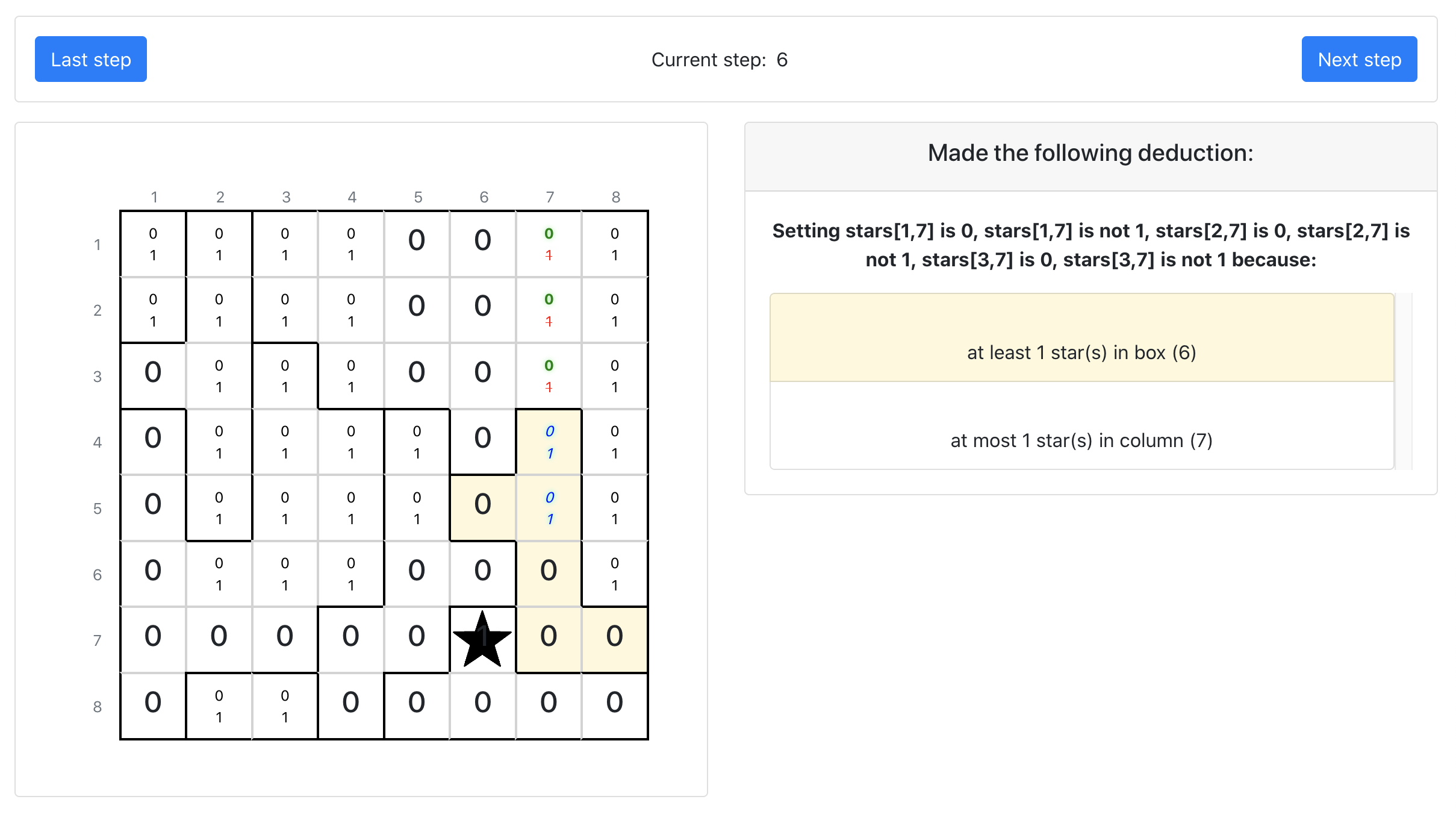}
    \end{center}
   \caption{An explanation produced by \demystify, as displayed in the visualiser interface.}
   \label{fig:demystifyvisualiser}
\end{figure}

\demystify includes a web-based visualiser that makes the explanations and functionalities offered by \demystify more understandable and useful to end users. This visualiser can display all the puzzles we consider in our experiments, showing step by step how to solve the puzzles. The interface displays any grid-based puzzle, and includes generic code to display typical parts of puzzles, including numbers around the edges of the grid, and highlighting regions of the grid, as can be seen in the Star Battle puzzle in \cref{fig:demystifyvisualiser}. These visualisations use the same input format as \demystify, so no extra code or formatting is required.

The \demystify Visualiser is a single-page web application \footnote{https://github.com/stacs-cp/Demystify-Visualiser} using React.js for a client-side interface, which communicates with a server-side API built using the Flask Python library. %

The user is able to upload their own \essence
input files, or select them from a predefined set, and then run \demystify in one of three modes. In the default mode \demystify will generate a series of MUSes which solve the entire puzzle. Secondly, there is a ``Choose MUSes manually" mode, where one step is sent to \demystify at a time, showing the visual interface between requests, and allowing the user to choose which of the smallest MUSes found is used, whenever \demystify finds multiple MUSes with the same smallest number of constraints. Finally, there is the ``Force choices" mode, which allows the user to force \demystify to produce an explanation for any literal, regardless of the associated MUS size. These interactive choices can be useful if the user wants to use \demystify to compare with their own solving techniques or solving techniques from other sources.

The visualiser shows the user one reasoning step at a time. 
Each step has a representation of the puzzle state on the left, showing the set of literals which have been removed. A list of the constraints, described in English, is shown along the right.
In order to clarify the parts of the puzzle relevant to any particular reason for a deduction, the literals involved in the corresponding MUS can be highlighted as part of a two-way mouse-over effect. 
When a literal is moused-over, all the constraints it is involved with are highlighted in the explanation list, and when an explanation is moused-over, all involved values on the grid are highlighted.

A screenshot of the visualiser with an explanation step is shown in Figure~\ref{fig:demystifyvisualiser}.

For the purposes of this tool, all puzzle states are reduced to a grid of cells. 
A solved cell contains a single bold value, while an unsolved cell contains a sub-grid of possible values. 
The values can be styled in a solving step to show how they are being used in a particular deduction: deduced true, deduced false, or involved in the reasoning. 

There are several choices we make when presenting MUSes to the user, which reduce information overload and allow the puzzles to be solved in fewer steps. These settings can all be configured, depending on the preferences of the user, or the particular puzzle being solved.

Firstly, if we find the smallest MUS at a given stage is size 1 (which refers to only a single constraint), we find all literals with MUSes of size 1 and remove all these literals in one step. Often puzzles such as Sudoku can at the start remove 30 or 40 literals with MUSes of size 1, so grouping these together greatly reduces the number of steps.

Secondly, as discussed in \cref{sec:musselect}, whenever we display a MUS we generate a list of the literals contained in it and handle all of these literals in a single step.

The final puzzle-specific choice is if \demystify is allowed to generate deductions of the form \(x \neq a\) for a variable \(x\) and value \(a\), or if it can only generate ``positive'' deductions for the form \(x = a\). In some puzzles, like Sudoku, it is common for players to deduce that some particular cell cannot take some value. In some other puzzles, such as Skyscrapers, it is common for users to only ever deduce the final value of a cell, and never try to remove single values. This configuration option can be chosen on a per-puzzle basis.

Because the time taken to produce an outputs for different puzzles can vary significantly, the application implements a simple Job-Queue pattern using a Redis database and the RQ library to manage requests for \demystify tasks and polling for output.

\section{MUS Algorithms}

As previously discussed, there are many existing algorithms for finding MUSes, in particular \cite{bogaerts2020step} use MUSes for solving puzzles. However, we found that these existing algorithms were not able to find MUSes for the hardest puzzles we considered. We therefore designed a randomised algorithm which we found gave small MUSes in practice, even on the hardest puzzles we considered. 

We began by considering the deletion-based algorithm of~\cite{10.1007/11814948_5} in~\cref{alg:mus}, which we give in \Cref{alg:origmus}. 
This algorithm, and all the other algorithms in this Section, require a SAT solver that provides a \textbf{FindUnsatCore} function. This function accepts a SAT 
problem $P$ and list of Boolean variables $L$, and return \textsc{Fail} if there is a solution where all variables in $L$ are \textsc{True}, or a subset of \(X\) which, if all assigned \textsc{True}, lead to an unsolvable problem. This set is not necessarily minimal (in particular, it is valid to always return all of $L$). Most modern SAT solvers provide this functionality.

\Cref{alg:origmus} finds a MUS by starting from an initial set $X$ and removing values one at a time, at each step checking if the problem is still unsolvable, and also using \textbf{FindUnsatCore} to try to reduce the current candidate MUS as far as possible.

\begin{algorithm}[t!]
\begin{algorithmic}[1]
\Procedure{DeleteMUS}{$P, X$}
    \State {$X$ = FindUnsatCore(\textbf{Shuffle}($X$))}
  \State {ToConsider = \textbf{ShuffledCopy}($X$)}
  \For {$c \in$ ToConsider}
    \If {$c \in X$}
        \State {newcore = \textbf{FindUnsatCore}($P$, $X - c$)}
        \If {newcore $\neq$ \textsc{Fail}}
         \State{core = newcore}
        \EndIf
    \EndIf
   \EndFor
   \State{ \Return {$X$}}
\EndProcedure
\end{algorithmic}
\caption{Deletion-Based MUS finding algorithm}
\label{alg:origmus}
\end{algorithm}

There are two main limitations of \Cref{alg:origmus} which limit its effectiveness for finding MUSes in puzzles:

\begin{enumerate}
    \item It can spend a long time finding MUSes which are very large ($>$ 300 variables), which are usually useless in practice (as in most cases there is some puzzle literal which will lead to a MUS of small size, usually less than 10).
    \item The MUSes it produces are often much larger than the smallest possible size. Changing the order in which elements of $X$ are considered does not improve this, because we found in practice the first few calls to \textbf{FindUnsatCore} lead to the a set which does not contain the smallest MUSes we want to find.
\end{enumerate}

We provide a new randomised algorithm which improves on both of these points. 

Firstly we present \textbf{LimitMUS}, a variant of \textbf{DeleteMUS}, in~\cref{alg:mus}. 
This algorithm accepts a problem $P$ and a set of variables $X$ (representing the constraints) and tests removing each element of $X$ in turn, checking the result is still unsolvable. It uses \textbf{FindUnsatCore} to reduce
the unsolvable subset of $X$ at each step. The only new feature is keeping track of the known required members of the MUS and stopping once $MaxSize$ members have been found and checking if they form a MUS, if not we need more values so we can return \textsc{Fail}.

\begin{algorithm}[t!]
\begin{algorithmic}[1]
\Procedure{LimitMUS}{$P, X,$ MaxSize}
    \State {$X$ = FindUnsatCore(\textbf{Shuffle}($X$))}
  \State {MusSize = 0} \Comment{Values known to be in MUS}
  \State {ToConsider = \textbf{ShuffledCopy}($X$)}
  \For {$c \in$ ToConsider}
    \If {$c \in X$}
        \State {core = \textbf{FindUnsatCore}($P$, $X - c$)}
        \If {core == \textsc{Fail}}
            \State {MusSize += 1}\Comment{$c$ must be in the core}
            \If {MusSize == MaxSize}\label{line:mussmall}
                \State{$X$ = $X[1..$MaxSize$]$}\label{line:stop}
                \If{FindUnsatCore($P$,$X$) == \textsc{Fail}} 
                \State\Return\textsc{Fail}
                	\Else{} 
                	\State\Return $X$
                \EndIf
            \EndIf
        \Else{} {$X$ = core}\label{line:secondfind}
        \EndIf
    \EndIf
   \EndFor
   \State{ \Return {$X$}}
\EndProcedure
\end{algorithmic}
\caption{Basic MUS finding algorithm}
\label{alg:mus}
\end{algorithm}

Our experiments demonstrate that while \textbf{LimitMUS} is significantly faster, it does not improve the variety of MUSes generated. The \textbf{ManyChop} algorithm, given in \cref{alg:muschop}, is intended to increase MUS variety while also improving performance.

\textbf{ManyChop} works by removing random subsets of $X$ and checking if the result is still unsatisfiable. The aim is that by considering many initial subsets of $X$, we increase the chances of seeing a variety of different MUSes -- this is demonstrated in our experiments.

\textbf{ManyChop} chooses a fixed-size proportion of $X$ to remove and keeps trying to remove that many elements of $X$, while checking for unsatisfiability

The mathematical idea behind the algorithm is that given a set \(X\), if we remove some
proportion \(p\) of the elements of \(X\), the probability any particular value is left behind is \(1-p\).
Therefore, the chance that a MUS of size \(n\) (assuming \(n\) is much smaller than \(|X|\) remains is approximately \((1-p)^n\). 

In our experiments, we choose \(p\) such that there is a probability of at least \(\frac{1}{10}\) of finding a MUS of size $MaxSize$ and then run the loop 20 times. 
We leave tuning the constants of this algorithm to future work.

One further advantage of \textbf{ManyChop} is that it is much more likely to remove large MUSes than small MUSes. If we are for example looking for a MUS \(M\) of size \(5\), then if we remove \(\frac{1}{4}\) of the elements in \(X\) there is a \(23.7\%\) chance that
\(M\) will not be removed. However, there is only a \(0.32\%\) chance a MUS of size \(15\) will not be removed. This suggests we should find smaller MUSes more often (when they exist) and informally we observe this in our experiments.

\begin{algorithm}
	\begin{algorithmic}[1]
	\Procedure{ManyChop}{$P,X,$ MaxSize}
	\State {$X$ = \textbf{Shuffle}($X$)}
	\State {step = min(\(\{n \in \mathbb{N}| (1-\frac{1}{2^n})^{\text{MaxSize}} \geq \frac{1}{10}\}\))}
        \State {frac = \(1-\frac{1}{2^{\text{step}}}\)}
	\For{$i \in \left[1..20\right]$}
        \State{check = $\textbf{Shuffle}(X)[1..|X|*\text{frac}]$}
		\If{\textbf{Solve}(check) == \textsc{False}}
            \State\Return{\textbf{LimitMUS}(check, MaxSize)}
		\EndIf
	\EndFor
	\State \Return \textsc{Fail}
	\EndProcedure
\end{algorithmic}
\caption{ManyChop Algorithm}
\label{alg:muschop}
\end{algorithm}

\textbf{ManyChop} takes a limit for the size of MUS to find. We use \cref{alg:allmus}  to find globally small MUSes. This accepts a SAT problem $P$, list of Boolean activators for the constraints $X$, problem literals $L$ (which represent the values these variables can take in the solution) and the number of times $n$ to search for each size of MUS (for our experiments we set $n=10$).

We search using iterative deepening, trying larger and larger sizes of MUS. In our experiments the loops on \cref{line:for1} are executed in parallel, distributing the calls to the MUS algorithm over
all available CPUs. We wait until \cref{line:ifret} to check if we have found a small enough MUS, rather than return as soon as a MUS of size $targetsize$ is found, as we may find many MUSes of the same size.

\begin{algorithm}[t!]
\begin{algorithmic}[1]
    \Procedure{FindGlobalMUS}{$P,X, L, n, \mathbf{musAlg}$}
  \State{SmallMUSd $= \textbf{FindSize1MUS}(P, X, L)$}
  \If{SmallMUSd $\neq$ \textsc{Fail}} 
  \State{\Return{SmallMUSd}}
  \EndIf
  \State{MUSd = \emph{dict}()}\Comment{Init as an empty dictionary}
  \State{smallest = $\infty$}
  \For{targetsize in $[1..|X|]$}
    \For{$r \in [1..n]$ and $l \in L$}\label{line:for1}
        \State{core = \textbf{MusAlg}($P + \{\neg l\}, X$, targetsize)}
        \If{core $\neq$ \textsc{Fail}}
            \If{($l \notin$ MUSd) \textbf{or} ($|\text{MUSd}[l]| > |\text{core}|)$}
                \State{MUSd$[l]$ = core}
                \State{smallest = min(smallest, $|\text{core}|$)}
            \EndIf
           
        \EndIf
    \EndFor
    \If{smallest $\leq$ targetsize}\label{line:ifret} 
    \State{\Return{MUSd}}
    \EndIf    
    \EndFor
\EndProcedure
\end{algorithmic}
\caption{Finding a globally smallest MUS}
\label{alg:allmus}
\end{algorithm}

\section{Experiments}

We consider two different methodologies to show that MUS generation via \textsc{FindGlobalMUS} in \demystify
lines up with how players solve puzzles. \demystify uses Glucose~\cite{DBLP:journals/ijait/AudemardS18} as the underlying SAT engine. Learned clauses are kept between calls to
the solver, which helps to speeds up the time taken to find many unsatisfiable sets~\cite{10.1007/978-3-642-39071-5_23}. 

Firstly, we compare against a selection of published tutorials. Secondly, we look at solving an entire puzzle where the player discusses their reasoning at each step. All experiments were run on a 6 Core 3.7GHz Intel i5-9600K with 16GB of RAM running Ubuntu 20.04 and Python 3.8.5.

\subsection{Tutorials}

To show that MUS generation lines up with how players solve puzzles, we compared the steps generated by \demystify to the tutorials for different puzzles, seeing in each case if the MUS highlighted the same constraints as those given by the tutorial.

For each step of each tutorial, we use \textbf{ManyChop} to get the smallest MUS for one of the deductions produced by that tutorial step. We do not use the globally smallest MUS, as in many cases there were smaller MUSes in different parts of the puzzle, unrelated to the logical rule the tutorial step was demonstrating.

In some cases, a MUS may only deduce one, or a subset, of the deductions described in a single tutorial step, as many tutorial steps describe a general idea and then apply it in many places. 
We define a successful match by the MUS when it correctly captures the reasoning for the single deduction we chose. 
Where tutorials show several connected steps we consider each step individually, rather than running \demystify to solve the whole puzzle.

There were two common issues we found with tutorials. In some cases the tutorial example had multiple answers, \demystify still works in this case, but will only deduce values which take the same values in all solutions. A small number of tutorial steps had no solutions, in this case, our algorithms do not work and we remove those instances.

We have taken instances from eight different online guides. For Sudoku, X-Sudoku and Jigsaw Sudoku we used~\citeS{Stuart2008SudokuWiki}.
The other two major sources for instances of techniques, for various puzzles, are~\citeA{conceptis}
and~\citeA{tectonic}. %
Some tutorials present named techniques with one or more example puzzles; in other cases, the explanations are spread over a step-by-step solving guide. 
\Cref{tab:SumRes} shows the total number of instances we extracted for each puzzle type, and how many times we matched the same required constraints as the tutorial.

For Binairo, Jigsaw Sudoku, Kakuro, Skyscrapers, Tents and Trees and X-Sudoku we matched all tutorial steps (\Cref{tab:SumRes}). On average for all puzzles, apart from classic Sudoku, we match 93\%. 
In some cases where \demystify produced a different MUS to the tutorial we believe it could be argued the MUS found by \demystify was simpler, but we strictly compare to the reasoning presented rather than apply our judgement as to which reasoning was simpler.

\begin{table}
\center
\begin{tabular}{ll|l|r|r|r|l}
\multicolumn{2}{l|}{Puzzle} & Source  & \#techniques & \multicolumn{2}{c}{matched}  \\
 & & & & \# & \% \\
\hline
\multicolumn{2}{l|}{Binairo} & \citeA{conceptis} & 7 & 7 & 100\%\\
\multicolumn{2}{l|}{} & \citeA{tectonic} & 6 & 6 & 100\%\\
\multicolumn{2}{l|}{Futoshiki} & \citeA{futoshikiorg} & 6 & 6 & 100\% \\
\multicolumn{2}{l|}{} & \citeA{tectonic} & 7 & 6 & 85\% \\
\multicolumn{2}{l|}{Jigsaw Sudoku} & \citeA{Stuart2008SudokuWiki} & 3 & 3 & 100\%\\
\multicolumn{2}{l|}{Kakuro} & \citeA{conceptis} & 11 & 7 & 64\% \\
\multicolumn{2}{l|}{} & \citeA{kakurocom} & 2 & 2 & 100\% \\
\multicolumn{2}{l|}{} & \citeA{kakurosorg} & 3 & 3 & 100\% \\
\multicolumn{2}{l|}{Skyscrapers} & \citeA{conceptis} & 14 & 12 & 85\% \\
\multicolumn{2}{l|}{Starbattle} & \citeA{krazydad} & 4 & 4 & 100\% \\
\multicolumn{2}{l|}{} & \citeA{tectonic} & 2 & 2 & 100\% \\
\multicolumn{2}{l|}{} & \citeA{rohanrao} & 11 & 14 & 78\% \\
\multicolumn{2}{l|}{} & \citeA{logicmasters} & 4 & 4 & 100\% \\

\multirow{2}{*}{Sudoku$\{$\hspace{-.2cm}}&Basic/Tough & \citeA{Stuart2008SudokuWiki} & 29 & 20 & 69\% \\
&Diabolical $\dagger$ & \citeA{Stuart2008SudokuWiki} & 29$\dagger$ & 12 & 41\% \\
\multicolumn{2}{l|}{Tents and Trees} & \citeA{tectonic} & 9 & 9 & 100\% \\
\multicolumn{2}{l|}{Thermometers} & \citeA{innoludic} & 2 & 2 & 100\% \\
\multicolumn{2}{l|}{} & \citeA{tectonic} & 5 & 4 & 80\% \\
\multicolumn{2}{l|}{X-Sudoku} & \citeA{Stuart2008SudokuWiki} & 3 & 3 & 100\%\\
\end{tabular}
\caption{Summary of the number of instances in guides, and how many \demystify matched. $\dagger$\emph{We exclude `Unique Rectangle' techniques, %
 which rely on the property that Sudokus have a unique answer. %
We use MUSes to check if a problem is unsolvable and not if it has a unique solution, so our technique does not  apply.}} %
\label{tab:SumRes}
\end{table}

Our results on the classic Sudoku puzzle are not as impressive as for the other puzzles. There are several reasons for this.
One is that we often find constraints which represent a different Sudoku technique to the one in the tutorial. For example, instead of the ``Naked Triples'' or ``Hidden Triple'' techniques we find ``Pointing Pairs'': the latter is sometimes considered to be an easier technique, e.g.~by  Sudoku Dragon's strategy guide~\cite{sudokudragon}.
A second reason is that Sudoku is exceptionally well-studied and many rules have been invented. Some of these `Diabolical'~\cite{Stuart2008SudokuWiki} techniques are required exceptionally rarely and many involve very large MUSes (up to 56 constraints), much larger than the MUSes in any of the other problems we looked at. We only accept these when we matched exactly and in many cases we found similar (and often smaller) but not identical reasoning.
We separate the ``Diabolical'' techniques in Table~\ref{tab:SumRes}, where we see significantly better performance on the `Basic' and `Tough' techniques.

Overall, we believe \cref{tab:SumRes} gives strong evidence for the validity of using MUSes for solving unseen puzzles. 
With no significant tuning (other than deciding how to represent \textit{AllDifferent} constraints) we have reproduced a significant number of the techniques from a varied set of puzzles.

\subsection{Miracle Sudoku}

One notable variant of Sudoku is the \emph{Miracle Sudoku}, designed by Mitchell Lee.
In the Miracle Sudoku the standard Sudoku rules apply, but cells separated by a king's move or knight's move in chess must have different values, and orthogonally adjacent cells cannot contain consecutive numbers.
A video showing an expert solving this puzzle achieved over one million views in under three months \cite{miracle} and resulted in mainstream press attention \cite{MiracleGuardian}. The puzzle, and final solution, are shown in \cref{fig:miracle}. To show our technique can solve entire puzzles, we compared \demystify against the solution given in \cite{miracle}.

\begin{figure}
\centering
	\begin{tikzpicture}[scale=.35]
	\begin{scope}
	\draw (0, 0) grid (9, 9);
	\draw[very thick, scale=3] (0, 0) grid (3, 3);
	\setcounter{row}{1}
	\setrow { }{ }{ }  { }{ }{ }  { }{ }{ }
	\setrow { }{ }{ }  { }{ }{ }  { }{ }{ }
	\setrow { }{ }{ }  { }{ }{ }  { }{ }{ }
	
	\setrow { }{ }{ }  { }{ }{ }  { }{ }{ }
	\setrow { }{ }{1}  { }{ }{ }  { }{ }{ }
	\setrow { }{ }{ }  { }{ }{ }  {2}{ }{ }
	
	\setrow { }{ }{ }  { }{ }{ }  { }{ }{ }
	\setrow { }{ }{ }  { }{ }{ }  { }{ }{ }
	\setrow { }{ }{ }  { }{ }{ }  { }{ }{ }
	
	\node[anchor=center] at (4.5, -0.9) {Initial State};
	\end{scope}
	
	\begin{scope}[xshift=12cm]
	\draw (0, 0) grid (9, 9);
	\draw[very thick, scale=3] (0, 0) grid (3, 3);
	\setcounter{row}{1}
	\setrow { }{ }{ }  { }{ }{ }  { }{ }{ }
\setrow { }{ }{ }  { }{ }{ }  { }{ }{ }
\setrow { }{ }{ }  { }{ }{ }  { }{ }{ }

\setrow { }{ }{ }  { }{ }{ }  { }{ }{ }
\setrow { }{ }{1}  { }{ }{ }  { }{ }{ }
\setrow { }{ }{ }  { }{ }{ }  {2}{ }{ }

\setrow { }{ }{ }  { }{ }{ }  { }{ }{ }
\setrow { }{ }{ }  { }{ }{ }  { }{ }{ }
\setrow { }{ }{ }  { }{ }{ }  { }{ }{ }
	\node[anchor=center] at (4.5, -0.9) {Completed};
	
	\begin{scope}[blue, font=\sffamily\slshape]
	\setcounter{row}{1}
\setrow {4}{8}{3}{7}{2}{6}{1}{5}{9}
\setrow {7}{2}{6}{1}{5}{9}{4}{8}{3}
\setrow {1}{5}{9}{4}{8}{3}{7}{2}{6}

\setrow {8}{3}{7}{2}{6}{1}{5}{9}{4}
\setrow {2}{6}{ }{5}{9}{4}{8}{3}{7}
\setrow {5}{9}{4}{8}{3}{7}{ }{6}{1}

\setrow {3}{7}{2}{6}{1}{5}{9}{4}{8}
\setrow {6}{1}{5}{9}{4}{8}{3}{7}{2}
\setrow {9}{4}{8}{3}{7}{2}{6}{1}{5}
	\end{scope}
	\end{scope}
	\end{tikzpicture}
	\caption{The Miracle Sudoku by Mitchell Lee.}
	\label{fig:miracle}
\end{figure}

We generated a full solution for the Miracle Sudoku.
The explanation contains  steps which involved MUSes of size 1 and a smaller number of more complex steps.
There were 49 steps where a non-unit MUS was used. These MUSes were size 3 or 4 and fell into one of the following categories. 
\begin{itemize}
\item A classic Sudoku technique such as pointing pairs.
\item A generalisation of this technique to Miracle Sudoku, for example where the only remaining possibilities for a 4 in a box are within a King's or Knight's move of another cell, that other cell cannot be 4.
\item What Anthony called ``dominoes'': if we know one of two adjacent cells must be 4 then neither can be either 3 or 5, by the consecutive numbers rule.
\item A similar but slightly more complex case of ``triominoes''. If we know that a 4 must occur in one of two non-adjacent cells, and both are adjacent to a third cell, the third cell cannot be 3 or 5.
\end{itemize}
Comparing these step-by-step with the YouTube video, we find that all the above techniques were used except the last. It is particularly striking that reasoning steps specialised to this variant, such as the use of dominoes, were discovered during solving both by Anthony and by our program.  Anthony never used the triomino technique above: in some cases, he used slightly more complicated reasoning steps in terms of the number of cells involved but ones which did not necessitate the discovery of the triomino reasoning step.  Both Anthony's and our explanation proceed similarly.

We also examined a second Miracle Sudoku solved by the YouTube channel  \cite{miracle-new}. Again, the broad outline of our explanation was very similar, this time being significantly more complex with more non-unit MUSes and MUSes of size 5. This correlates with Anthony's observation that this puzzle seemed harder. Our explanation was slightly worse than the human's: there were 4 reasoning steps which involved a choice of several MUSes of size 5 and the ones our solver used were harder to explain than the ones the human solver used.

While this section only reports on a single type of Sudoku variant, it is promising that our techniques could find broadly similar explanations to a human expert.

\subsection{Performance Comparison}

While performance is not a primary concern in this paper, we performed one small experiment to compare the performance of our algorithms. 
This experiment demonstrates the need for a SAT solver which has the functionality of \texttt{FindUnsatCore}, and also shows the wider variety of MUSes that \texttt{ManyChop} finds.

The algorithms solved six Sudokus from August 20th 2020 to August 30th 2020 from the LA Times which require at least one MUS of size greater than 1. 
To ensure the algorithms were forced to consider the same steps, we first solved each Sudoku with the \texttt{LimitMUS} technique, and then use the same sequence of choices for all the other algorithms. 

Our results are presented in \cref{tab:latimes}. 
We ran our algorithms with \texttt{FindUnsatCore} always returning its input when the problem is unsolvable instead of a subset (\texttt{-Core}), and with no limit on the size of the MUS to be found (\texttt{-Limit}). 
We measured, each time the algorithms had to find a MUS of size greater than 1, both the number of candidates with MUS of the smallest size found by any algorithm and the total number of distinct smallest size MUSes found for all candidates. 

We observe the \texttt{-Core} variants are orders of magnitudes slower, so a good implementation of \texttt{FindUnsatCores} is vital for MUS finding. The \texttt{Basic} algorithm, while fast, produces the fewest different smallest MUSes.  The \texttt{ManyChop} algorithm performs faster and also produces a much greater number of smallest MUSes.

\begin{table}
\centering
	\begin{tabular}{l|r|r|r|r}
	Technique & \#& Time & Choices & MUSes \\
	\hline
\textbf{Basic}            & 6          & 92         & 80         & 125        \\
\textbf{\quad -Limit} 		  & 6          & 140        & 82         & 136        \\
\textbf{\quad -Core}  	  & 4          & -      & 101        & 178        \\
\textbf{\quad -Limit, -Core} 	  & 3          & -       & 70         & 129        \\
\hline
\textbf{ManyChop}         & 6          & 49         & 124        & 312        \\
\textbf{\quad -Core}     & 6          & 13086      & 124        & 323        \\
\end{tabular}
\caption{Solving 6 LA Times Sudokus to completion. Solvers run with no FindUnsatCores (Core) and no MUS size limit (Limit) where appropriate. Times in CPU mins, solvers given 9600 CPU mins to solve each Sudoku. \# - Total solved, Choices - total number of candidates with smallest MUS size, MUSes - total number of distinct MUSes found for all these candidates.}
\label{tab:latimes}
\end{table}

\section{Conclusion and Future Work}

In this paper, we have present a new algorithm for efficiently finding small MUSes. We demonstrate its usefulness and generality by producing descriptions of steps for many pen and  paper puzzles. 

Our experiments demonstrate that MUSes align very closely with pre-existing guides on how human players 
decide how to solve these puzzles. 
This work, along with earlier work on Logic Grid %
Puzzles~\cite{bogaerts2020step}, provides strong evidence that MUSes are a powerful, natural, and
generic method of explaining how to solve puzzles in a human-like way.

MUSes do not perfectly match up with how guides tell us puzzles should be solved. In future work we plan to perform experiments to see how players solve puzzles and investigate where players deviate from choosing the smallest MUS.

While \demystify is a generic system, there are some types of puzzles where at present MUSes to not currently produce good quality explanations, in particular puzzles which require parts of the puzzle contain cycles, or are globally connected. This is because existing constraint models for cycle and globally connected constraints do not break up into small pieces which can then produce useful MUSes. In future we will plan to investigate how to model such problems in a way that \demystify can produce useful explanations.

There are also puzzles such as Killer Sudoku where an interplay of global, local and implicit constraints in solving techniques is proving difficult to model such that MUS explanations correlate with how people solve these puzzles.

While MUSes show the users which parts of the puzzle to consider, more complicated MUSes may need splitting into smaller substeps. We want to investigate how to explain an individual MUS -- which requires first understanding how human puzzles players divide a single MUS into multiple reasoning steps. %

\bibliography{main}
\bibliographystyle{theapaurl}

\end{document}